\documentclass[sigconf]{aamas} 

\usepackage{balance} 
\usepackage{subcaption}

\usepackage{amsmath}
\usepackage{relsize}
\usepackage{algorithmic}
\usepackage[ruled,vlined,linesnumbered,noresetcount]{algorithm2e}
\usepackage{mathrsfs}
\usepackage[switch]{lineno}
\usepackage{caption}
\usepackage{subcaption}

\DeclareMathOperator*{\argmax}{arg\,max}

\newcommand{\expectedvalue}{\mathbb{E}_{\policy}}
\newcommand{\true}{\mathit{true}}
\newcommand{\false}{\mathit{false}}

\newcommand{\task}{T}
\newcommand{\fluents}{F}

\newcommand{\operators}{O}
\newcommand{\initfluentstate}{\sigma_0}
\newcommand{\goal}{\partialfluentstate_g}
\newcommand{\subgoal}{\partialfluentstate_{sg}}

\newcommand{\fluent}{f}

\newcommand{\fluentstate}{\sigma}
\newcommand{\partialfluentstate}{\tilde{\sigma}}
\newcommand{\operator}{o}
\newcommand{\partialoperator}{\tilde{o}}
\newcommand{\plan}{\pi_T}
\newcommand{\precon}{pre}
\newcommand{\effect}{\mathit{eff}}

\newcommand{\static}{static}

\newcommand{\allliterals}{\mathcal{L}(\fluents)}
\newcommand{\fluenttransition}{\delta(\partialfluentstate, \operator)}

\newcommand{\regressoperator}[2]{\delta^{-1}(#1, #2)}
\newcommand{\applyfunction}{\delta}
\newcommand{\regressfunction}{\delta^{-1}}

\newcommand{\reachable}{\Delta_{\partialfluentstate}}

\newcommand{\restrict}{restrict}

\newcommand{\planfluentstates}{\Sigma_{plan}}
\newcommand{\reachfluentstates}{\Sigma_{reach}}

\newcommand{\mdp}{M}
\newcommand{\mstates}{S}
\newcommand{\actions}{A}
\newcommand{\rewardfunction}{r}
\newcommand{\transitionprob}{p}
\newcommand{\discountfactor}{\gamma}
\newcommand{\mstate}{s}
\newcommand{\initialstatedist}{\iota}
\newcommand{\initialstate}{\mstate_0}
\newcommand{\action}{a}
\newcommand{\valuefunction}{v_{\policy(\mstate)}}
\newcommand{\policy}{\pi_M}
\newcommand{\policies}{\Pi_M}

\newcommand{\qfunction}{q(\mstate, \action)}
\newcommand{\qfunctionopt}{q^*(\mstate, \action)}
\newcommand{\qfunctionnext}{q(\newstate, \newaction)}
\newcommand{\learningrate}{\alpha}

\newcommand{\newstate}{\mstate^\prime}
\newcommand{\newaction}{\action^\prime}
\newcommand{\onlinelearner}{\ell_{expl}}
\newcommand{\offlinelearner}{\ell}
\newcommand{\learners}{L}
\newcommand{\threshold}{\tau}

\newcommand{\exec}{x}
\newcommand{\execinit}{I_x}
\newcommand{\execpolicy}{\pi_x}
\newcommand{\execterm}{\beta_x}
\newcommand{\execs}{X}

\newcommand{\execstar}{x^\star}
\newcommand{\execinitstar}{I_{x^\star}}
\newcommand{\execpolicystar}{\pi_{x^\star}}
\newcommand{\exectermstar}{\beta_{x^\star}}

\newcommand{\execinitfor}[1]{I_{#1}}

\newcommand{\exectermfor}[1]{\beta_{#1}}

\newcommand{\explorationpolicy}{\pi_{expl}}

\newcommand{\symdp}{\mathcal{T}}
\newcommand{\detector}{d}
\newcommand{\executor}{e}
\newcommand{\macgyver}{\tilde{\symdp}}

\newcommand{\beenfluentstates}{\Sigma_{been}}
\newcommand{\abovethresholdfluentstates}{\Sigma_{>\tau}}
\newcommand{\node}{node}
\newcommand{\common}{\partialfluentstate_{common}}
\newcommand{\anotherfluentstate}{\fluentstate^\prime}

\newcommand{\lflag}{\mathit{impasse}}

\newcommand{\solve}{\textbf{\textsc{solve}}}
\newcommand{\learn}{\textbf{\textsc{learn}}}
\newcommand{\genprecon}{\textbf{\textsc{gen-precon}}}
\newcommand{\execute}{\textsc{execute}}

\newcommand{\owfs}{\textsc{owfs}}

\setcopyright{ifaamas}
\acmConference[AAMAS '21]{Proc.\@ of the 20th International Conference on Autonomous Agents and Multiagent Systems (AAMAS 2021)}{May 3--7, 2021}{London, UK}{U.~Endriss, A.~Now\'{e}, F.~Dignum, A.~Lomuscio (eds.)}
\copyrightyear{2021}
\acmYear{2021}
\acmDOI{}
\acmPrice{}
\acmISBN{}

\acmSubmissionID{37}

\title{SPOTTER: Extending Symbolic Planning Operators through Targeted
Reinforcement Learning}

\author{Vasanth Sarathy}
\authornote{The first two authors contributed equally.}
\affiliation{
  \institution{Smart Information Flow Technologies}
  \city{Lexintgon}
	\state{MA}}
\email{vsarathy@sift.net}

\author{Daniel Kasenberg}
\authornotemark[1]
\affiliation{
  \institution{Tufts University.}
  \city{Medford}
  \state{MA}}
\email{dmk@cs.tufts.edu}

\author{Shivam Goel}
\affiliation{
	\institution{Tufts University.}
	\city{Medford}
	\state{MA}}
\email{shivam.goel@tufts.edu}

\author{Jivko Sinapov}
\affiliation{
	\institution{Tufts University.}
	\city{Medford}
	\state{MA}}
\email{jivko.sinapov@tufts.edu}

\author{Matthias Scheutz}
\affiliation{
	\institution{Tufts University.}
	\city{Medford}
	\state{MA}}
\email{matthias.scheutz@tufts.edu}

\begin{abstract}
	Symbolic planning models allow decision-making agents to sequence actions in
arbitrary ways to achieve a variety of goals in dynamic domains. However, they
are typically handcrafted and tend to require precise formulations that are not
robust to human error. Reinforcement learning (RL) approaches do not require
such models, and instead learn domain dynamics by exploring the environment and
collecting rewards. However, RL approaches tend to require millions of episodes
of experience and often learn policies that are not easily transferable to
other tasks. In this paper, we address one aspect of the open problem of
integrating these approaches: how can decision-making agents resolve
discrepancies in their symbolic planning models while attempting to accomplish
goals? We propose an integrated framework named SPOTTER that uses RL to augment
and support (``spot'') a planning agent by discovering new operators needed
by the agent to accomplish goals that are initially unreachable for the agent.
SPOTTER outperforms pure-RL approaches while also discovering transferable
symbolic knowledge and does not require supervision, successful plan traces or
any \emph{a priori} knowledge about the missing planning operator.

\end{abstract}


\newcommand{\BibTeX}{\rm B\kern-.05em{\sc i\kern-.025em b}\kern-.08em\TeX}

\begin{document}


\pagestyle{fancy}
\fancyhead{}


\maketitle 


\section{Introduction}

Symbolic planning approaches focus on synthesizing a sequence of operators capable of
achieving a desired goal \cite{ghallab2016automated}. These approaches rely on
an accurate high-level symbolic description of the dynamics of the environment.
Such a description affords these approaches the benefit of generalizability and
abstraction (the model can be used to complete a variety of tasks), available
human knowledge, and interpretability. However, the models are often
handcrafted, difficult to design and implement, and require precise
formulations that can be sensitive to human error. Reinforcement learning (RL)
approaches do not assume the existence of such a domain model, and instead
attempt to learn suitable models or control policies by trial-and-error
interactions in the environment \cite{sutton2018reinforcement}. However, RL
approaches tend to require a substantial amount of training  in moderately
complex environments. Moreover, it has been difficult to learn abstractions to
the level of those used in symbolic planning approaches through low-level
reinforcement-based exploration.  Integrating RL and symbolic planning is
highly desirable, enabling autonomous agents that are robust, resilient and
resourceful.

Among the challenges in integrating symbolic planning and RL, we focus here on
the problem of how partially specified symbolic models can be extended during
task performance. Many real-world robotic and autonomous systems already have
preexisting models and programmatic implementations of action hierarchies.
These systems are robust to many anticipated situations. We are interested in
how these systems can adapt to unanticipated situations, autonomously and with
no supervision.

In this
paper, we present SPOTTER (Synthesizing Planning Operators through Targeted
Exploration and Reinforcement). Unlike other approaches to action model
learning, SPOTTER does not have access to successful symbolic traces. Unlike
other approaches to learning low-level implementation of symbolic operators,
SPOTTER does not know \emph{a priori} what operators to learn. Unlike other
approaches which use symbolic knowledge to guide exploration, SPOTTER does not
assume the existence of partial plans.  

We focus on the case where the agent is faced with a symbolic goal but has
neither the necessary symbolic operator to be able to synthesize a plan nor its
low-level controller implementation. The agent must invent both. SPOTTER
leverages the idea that the agent can proceed by planning alone unless there is a 
discrepancy between its model of the
environment and the environment's dynamics. In our approach, the agent attempts
to plan to the goal. When no plan is found, the agent explores its
environment using an online explorer looking to reach any state from which it
can plan to the goal.  When such a state is reached, the agent spawns offline
``subgoal'' RL learners to learn policies which can consistently achieve those
conditions. As the exploration agent acts, the subgoal learners learn from its
trajectory in parallel. The subgoal learners regularly attempt to generate
symbolic preconditions from which their candidate operators have high value; if
such preconditions can be found, the operator is added with those preconditions
into the planning domain. We evaluate the approach with experiments in a
gridworld environment in which the agent solves three puzzles involving
unlocking doors and moving objects. 

In this paper, our contributions are as follows: a framework for integrated RL and symbolic planning; algorithms for solving problems in finite,
deterministic domains in which the agent has partial domain knowledge and must reach a seemingly
unreachable goal state; and experimental results showing substantial improvements over baseline
approaches in terms of cumulative reward, rate of learning and
transferable knowledge learned.  

\subsection{Running example: GridWorld}

\begin{figure}[b]
	\centering
	\includegraphics[width=1.7in]{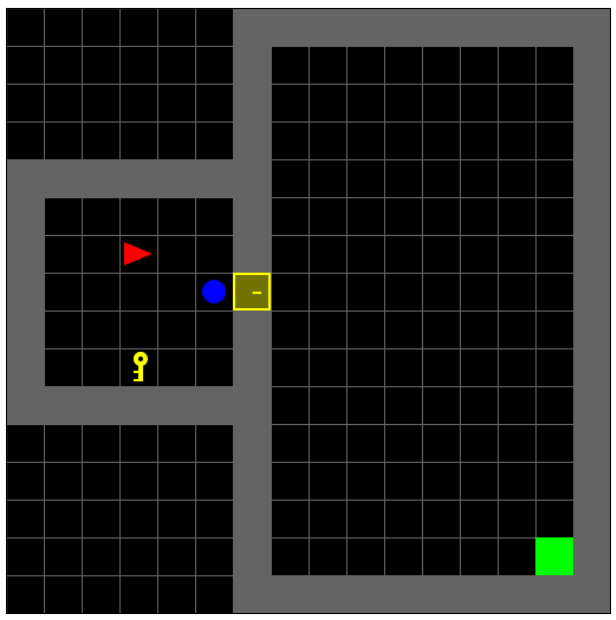}
	\caption{The agent's (red triangle) goal is to unlock the door
		(yellow square). SPOTTER can learn how to move the blue ball
		out of the way. The learned representation is
		symbolic and can be used to plan (without additional learning) to achieve
		different goals like reaching the green square.}
	\label{fig:screenshot}
\end{figure}

Throughout this paper, we will use as a running example a GridWorld puzzle (Figure~\ref{fig:screenshot}) which an agent must unlock a door which is blocked by a ball. The agent's planning domain abstracts out the notion of specific grid cells, and so all navigation is framed in terms of going to specific objects. Under this abstraction, no action sequence allows the agent to ``move the ball out of the way''. Planning actions are implemented as handcrafted
programs that navigate the puzzle and execute low-level actions (up,
down, turn left/right, pickup). Because the door is blocked, the agent cannot generate a symbolic
plan to solve the problem; it must synthesize new symbolic actions or operators. An agent using SPOTTER performs RL on low-level actions to learn to reach states from which a plan to the goal exists, and thus can learn a symbolic operator corresponding to ``moving the ball out of the way''.

\section{Background}

In this section, we provide a background of relevant concepts in planning and
learning. 

\subsection{Open-World Symbolic Planning}
\label{sec:planning}

We formalize the planning task as an open-world variant of propositional
STRIPS \cite{fikes1971strips}, $\task = \langle \fluents, \operators, \initfluentstate, \goal
\rangle$. We consider $\fluents$ (fluents) to be the set of propositional state
variables $\fluent$. A fluent state $\fluentstate$ is a complete assignment of
values to all fluents in $\fluents$. That is, $|\fluentstate| = |\fluents|$,
and $\fluentstate$ includes positive
literals ($\fluent$) and negative literals ($\lnot \fluent$).
$\initfluentstate$ represents the initial fluent state. We assume full
observability of the initial fluent state. We can define a partial fluent state $\partialfluentstate$ to refer to a partial
assignment of values to the fluents $\fluents$.  The goal condition is represented as a partial fluent state $\goal$.  We define $\allliterals$ to be the set of all partial fluent states with fluents in $\fluents$.

 The operators under this formalism are open-world. A partial
 planning operator can be defined as $\operator = \langle \precon(\operator),
 \effect(\operator),  \static(\operator) \rangle$.  $\precon(\operator) \in
 \allliterals$ are the preconditions, and $\effect(\operator) \in \allliterals$ are
 the effects, $\static(\operator) \subseteq F$ are those fluents whose
 values are \emph{known} not to change during the execution of the operator.  An
 operator $\operator$ is \emph{applicable} in a partial fluent state
 $\partialfluentstate$ if $\precon(\operator) \subseteq \partialfluentstate$.
 The result of executing an operator $\operator$ from a partial fluent state
 $\partialfluentstate$ is given by the \emph{successor} function
 $\fluenttransition = \effect(\operator) \cup \restrict(\partialfluentstate, \static(\operator))$, where $\restrict(\partialfluentstate, \fluents')$ is the partial fluent state consisting only of $\partialfluentstate$'s values for fluents in $\fluents'$. The set of all partial fluent states defined through
 repeated application of the successor function beginning at
 $\partialfluentstate$ provides the set of reachable partial fluent states
 $\reachable$. A complete operator is an operator $\partialoperator$ where all
 the fluents are fully accounted for, namely, $\forall \fluent \in \fluents,
\fluent \in \effect^+(\partialoperator) \cup \effect^-(\partialoperator) \cup \static(\partialoperator)$,  where $\effect^+(\operator) = \{ f\in F: f \in \effect(\operator) \}$ and $\effect^-(\operator)=\{f \in F: \neg f \in \effect(\operator)\}$. We assume all operators $\operator$ satisfy $\effect(\operator) \backslash \precon(\operator) \neq \emptyset$; these are the only operators useful for our purposes.\footnote{We also assume  $(\effect^+(\operator) \cup \effect^-(\operator)) \cap \static(\operator) = \emptyset$; i.e. that postconditions and static variables do not conflict.} A plan $\plan$ is a sequence of operators $\langle \operator_1, \ldots,
\operator_n\rangle$. A plan $\plan$ is executable in state $\initfluentstate$ if, for all $i\in \{1, \cdots, n\}$, $\precon(\operator_i) \subseteq \partialfluentstate_{i-1}$ where $\partialfluentstate_i=\delta(\partialfluentstate_{i-1}, \operator_i)$. A plan $\plan$ is said to solve the task $\task$ if executing
$\plan$ from $\initfluentstate$ induces a
trajectory $\langle \initfluentstate, \operator_1, \partialfluentstate_1,
\ldots, \operator_n, \partialfluentstate_n \rangle$ that reaches the goal
state, namely $\goal \subseteq \partialfluentstate_n$.

An open-world forward search (OWFS) is a 
breadth-first plan search procedure where each node is a partial fluent state $\partialfluentstate$. The successor
relationships and applicability of $\operators$ are specified as defined above
and used to generate successor nodes in the search space.
A plan is synthesized once $\goal$ has been reached. If $\initfluentstate$ is a complete state and the operators are complete operators, then each
node in the search tree will be complete fluent states, as well. 

We also define the notion of a ``relevant''
operator and ``regressor'' nodes, as used in backward planning search \cite{ghallab2016automated}. For a partial fluent state $\partialfluentstate$ with fluents
$\fluent_i$ and their valuations $c_i \in
\{True, False\}$, operator $\operator$ is \emph{relevant} at partial fluent state $\partialfluentstate$
when:

\begin{enumerate}
	\item $\restrict(\partialfluentstate \backslash \effect(\operator), \static(\operator)) = \partialfluentstate \backslash \effect(\operator)$ and 
	\item $\partialfluentstate \supseteq \effect(\operator) \cup \restrict(\precon(\operator), \static(\operator))$
\end{enumerate}

When a relevant operator $\operator$ is found for a particular
$\partialfluentstate$, it can be \emph{regressed} to generate a partial fluent state 
$\regressoperator{\partialfluentstate}{\operator}= \precon(\operator) \cup (\partialfluentstate  \backslash\effect(\operator))$.\footnote{This is a slight abuse of notation because $\regressfunction$ is not the function inverse of $\applyfunction$, but they can be thought of as inverse in the sense described in this paragraph.} Regression is the ``inverse'' of operator application in that if applying any operator sequence $\langle \operator_1, \ldots, \operator_n \rangle$ yields final state $\partialfluentstate'$, if we let $\partialfluentstate''=\regressoperator{\ldots(\regressoperator{\partialfluentstate}{\operator_n}\ldots)}{\operator_1}$, then $\partialfluentstate'' \subseteq \partialfluentstate$.  In particular, $\partialfluentstate''$ is the \textit{minimal} partial fluent state from which application of $\langle \operator_1, \cdots, \operator_n \rangle$ results in $\partialfluentstate'$.

\subsection{Reinforcement Learning}

We formalize the environment in which the agent acts as a Markov Decision
Process (MDP) $\mdp = \langle \mstates, \actions, \transitionprob, \rewardfunction, \initialstatedist, \discountfactor \rangle$, where $\mstates$ is the set of
states, $\actions$ is the set of actions, $\transitionprob$ is the probability distribution
$p(s_{t+1} \mid s_t, a_t)$, $\rewardfunction: \mstates \times
\actions \times \mstates \rightarrow \mathbb{R}$ is a reward function, $\initialstatedist$ is a probability distribution over initial states, and $\discountfactor \in (0, 1]$ \cite{sutton2018reinforcement}. A policy for
$\mdp$ is defined as the probability distribution $\policy(\action \mid \mstate)$ that establishes the probability of an agent taking an action
$\action$ given that it is in the current state $\mstate$.  We define the set of all such policies in $\mdp$ as $\policies$.  We let $\mstates_0 = \{ \mstate \in \mstates: \initialstatedist(\mstate) > 0\}$. An RL problem typically consists of finding an optimal policy $\policy^* \in \policies$ that
maximizes the expected discounted future rewards obtained from $\mstate \in
\mstates$:
$$ \policy^* = \argmax_{\policy} \mathlarger{\sum}_{\mstate \in \mstates}
\valuefunction, $$

\noindent where $\valuefunction$ is the value function and captures the expected
discounted future rewards obtained when starting at state $\mstate$ and
selecting actions according to the policy $\policy$:
$$\valuefunction = \expectedvalue \left[  \mathlarger{\sum}_{t=0}^{\infty} \discountfactor^t \rewardfunction_t
	\mid \initialstate = \mstate
\right]. $$

At each time step, the agent executes an action $\action$ and the environment
returns the next state $\mstate^\prime \in \mstates$ (sampled from
\transitionprob) and an immediate reward $\rewardfunction$. The experience is
then used by the agent to learn and improve its current policy $\policy$.

Q-learning \cite{watkins1992q} is one learning technique in which an agent uses experiences
to estimate the optimal Q-function $\qfunctionopt$ for every state $\mstate \in
\mstates$ and $\action \in \actions$, where $\qfunctionopt$ is the expected
discounted sum of future rewards received by performing action $\action$ in
state $\mstate$. The Q-function is updated as follows:
$$  \qfunction \gets \qfunction + \learningrate \left[ \left( \rewardfunction
		+ \discountfactor \max_{\action^\prime \in \actions} \qfunctionnext - \qfunction
\right) \right],$$
where $\learningrate \in (0, 1]$ is the learning rate. The Q-learner can explore the environment, e.g., by following an
$\epsilon$-greedy policy, in which the agent selects a random action with probability
$\epsilon$ and otherwise follows an action with the largest $\qfunction$.

\section{Proposed SPOTTER Framework}

We begin by introducing a framework for integrating the planning and learning
formulations. We  define an integrated planning task that enables us to ground
symbolic fluents and operators in an MDP, specify goals symbolically, and
realize action hierarchies.

We define an \textit{executor} for a given MDP $\mdp=\langle \mstates, \actions, \transitionprob, \rewardfunction,  \initialstatedist, \discountfactor \rangle$ as a triple $\exec=\langle \execinit, \execpolicy, \execterm \rangle$ where $\execinit \subseteq \mstates$ is an initiation set, $\execpolicy(\action|\mstate_{init}, \mstate)$ is the probability of performing $\action$ given that the executor initialized at state $\mstate_{init}$ and the current state is $\mstate$, and $\execterm(\mstate_{init}, \mstate)$ expresses the probability of terminating $\exec$ at $\mstate$ given that $\exec$ was initialized at $\mstate_{init}$.  \footnote{An executor is simply an option \cite{sutton1999between}
	where the policy and termination condition depend on where it was initialized.}  We define $\execs_\mdp$ as the set of executors for $\mdp$.
\begin{definition} (Integrated Planning Task)
	We can formally\\  define an \emph{Integrated Planning Task} (IPT) as $\symdp =
	\langle \task, \mdp, \detector, \executor
	\rangle $ where $\task = \langle \fluents, \operators, \initfluentstate, \goal
	\rangle$ is an open-world STRIPS task, $\mdp=\langle \mstates, \actions,
	\transitionprob, \rewardfunction,  \initialstatedist, \discountfactor \rangle$ is an MDP,  a
	detector function $\detector : \mstates \mapsto \allliterals$ determines a
	fluent state for a given MDP state, and an executor function $\executor :
	\operators \mapsto \execs_\mdp$ that determines a mapping between an operator and an executor in the underlying MDP.
\end{definition}
	For the purposes of this paper, we assume that for each operator $\operator \in \operators$, $\executor(\operator)$ is \textit{accurate} to $\operator$; that is, for every $\operator \in \operators$, $\execinitfor{\executor(\operator)} \supseteq \{ \mstate \in \mstates: \detector(\mstate) \supseteq \precon(\operator) \}$ and
	\begin{equation*}
	\exectermfor{\executor(\operator)}(\mstate_{init}, \mstate) = \begin{cases}
	1 &\textrm{if }\detector(\mstate) \supseteq \effect(\operator)  \\ & \quad \cup \restrict(\detector(\mstate_{init}), \static(\operator)) \\
	0 & \textrm{otherwise}.
	\end{cases}
	\end{equation*}
		
	The objective of this task is to find an executor $\execstar\in \execs_\mdp$ such that $\exectermstar(\mstate_{init}, \mstate) = 1$ if and only if $\detector(\mstate) \supseteq \goal$, and $\execinitstar \supseteq \mstates_0$, and $\execstar$ terminates in finite time.
	
	A \textit{solution} to a particular IPT $\symdp$ is an executor $\execstar\in \execs_\mdp$ having the properties defined above.
	
	A \textit{planning solution} to a particular IPT $\symdp$  is a mapping $\plan:\mstates_0 \rightarrow \operators^*$ such that for every $\mstate_0 \in \mstates_0$, 
$\plan(\mstate_0) = \langle \operator_1, \ldots, \operator_n \rangle$ is executable at $\mstate_0$ and achieves goal state $\goal$.  Assuming all operators are accurate, executing in the MDP $\mdp$ the corresponding executors $\executor(\operator_1), \ldots, \executor(\operator_n)$ in sequence will yield a final state $\mstate$ such that $\detector(\mstate) \supseteq \goal$ as desired. 

An IPT $\symdp$ is said to be solvable if a
solution exists. It is said to be plannable if a planning solution exists. 

\subsection{The Operator Discovery Problem}

As we noted earlier, symbolic planning domains can be sensitive to human
errors. One common error is when the domain is missing an 
operator, which then prevents the agent from synthesizing a plan that requires
such an operator. We define a stretch-Integrated Planning Task,
stretch-IPT\footnote{akin to ``stretch goals'' in business productivity}, that captures
difficult but achievable goals -- those for which missing operators must be
discovered.

\begin{definition}
	(Stretch-IPT). A Stretch-IPT $\macgyver$ is an
	IPT $\symdp$ for which a solution exists, but a planning solution does not.
\end{definition}
Sarathy et al.
considered something similar in a purely symbolic planning context as MacGyver Problems
\cite{sarathy2018macgyver}; here we extend these ideas to integrated symbolic-RL domains. A planning solution is desirable because a plannable task
affords the agent robustness to variations in goal descriptions and a certain
degree of generality and ability to transfer decision-making capabilities
across tasks. We are interested in turning a stretch-IPT into an
plannable IPT, and specifically study how an agent can automatically extend its
task description and executor function to invent new operators and their
implementations.

\begin{definition}
	(Operator Discovery Problem). Given a stretch-IPT $\macgyver = \langle \task, \mdp, \detector,
\executor  \rangle $ with $\task=\langle \fluents, \operators, \initfluentstate, \goal \rangle$, construct a set of operators $\operators^\prime = \{\operator'_1, \cdots, \operator'_m\}$ and their executors $\exec_{\operator'_1}, \cdots, \exec_{\operator'_m} \in \execs$ such that the IPT $\langle \task', \mdp, \detector, \executor' \rangle$ is plannable, with $\task' = \langle \fluents, \operators \cup \operators', \initfluentstate, \goal \rangle$ and the executor function \begin{equation*} \executor'(\operator) = \begin{cases} \executor(\operator) & \textrm{if }\operator \in \operators \\ \exec_{\operator} &\textrm{if }\operator \in \operators' \end{cases}.\end{equation*}
\end{definition}


In the rest of the section, we will outline an approach for solving the
operator discovery problem. 


\subsubsection{The Operator Discovery Problem in GridWorld}

In the example GridWorld puzzle, the SPOTTER agent is equipped with a high-level planning domain specified in an
open-world extension of PDDL that can be grounded down into an open-world
STRIPS problem. This domain is an abstraction of the environment
which ignores the absolute positions of objects.  In particular, the core
movement actions $\mathit{forward}$, $\mathit{turnLeft}$, and $turnRight$ in the MDP are absent
from the planning domain, which navigates in terms of objects using, e.g., the
operator $goToObj(agent, object)$, with
preconditions $\neg holding(agent, object)$, $\neg blocked(object)$, and $inRoom(agent, object)$, and with effect  $nextToFacing(agent, object)$. All initial operators are assumed to satisfy the closed-world assumptions, except that putting down an object ($putDown(agent, object)$) leaves unknown
whether at the conclusion of the action, some other object (e.g., a door) will
be blocked.  Each operator has a corresponding hand-coded executor. The goal here is $\goal= \{ open(door)\}$, which is not achievable using planning alone from the initial state.


\subsection{Planning and Execution}

Our overall agent algorithm is given by Algorithm
\ref{algo:overall}. We
consider the agent to be done when it has established that the input IPT
$\symdp$ is a plannable IPT. This is true once the agent is able to
find a complete solution to the IPT through planning and execution, alone, and
without requiring any learning. Algorithm \ref{algo:overall} shows that the agent
repeatedly learns in the environment until it has completed a run in which
there were no planning impasses, i.e., situations where no plan was found through a
forward search in the symbolic space.  The set of learners $\learners$ is maintained between runs of $\solve$.

%
\begin{algorithm}[b]\centering
	\small
		\caption{\textbf{\textsc{spotter}}($\symdp$)}
		\begin{algorithmic}[1]
			\REQUIRE {$\symdp$: Integrated Planning Task}
			\STATE {$\lflag \gets \true$}
			\STATE{$\learners \gets \{ \onlinelearner \}$} 
			\WHILE {$\lflag$}
			\STATE {$\initialstate\sim \task.\mdp.\initialstatedist(\cdot)$}
			\STATE {$\lflag, \learners \gets$ \solve($\symdp$, $\initialstate$, $\emptyset$, $\emptyset$, $\threshold$, $\false$, $\learners$))}
			\ENDWHILE
			\RETURN $\symdp$
		\end{algorithmic}
		\label{algo:overall}
\end{algorithm}


Two sets of fluent states will be important in Algorithms \ref{algo:solve}-\ref{algo:genprecon}: $\reachfluentstates$ and $\planfluentstates$. $\reachfluentstates$ contains all states known to be reachable from the initial state via planning. $\planfluentstates$ contains all states from which a plan to the goal $\goal$ is known.

Algorithm \ref{algo:solve} (\solve) begins with the agent
performing an OWFS (open-world forward search) of the symbolic space specified
by the task. If a plan is found (line 7), the agent attempts to execute the
plan (line 9).  If unexpected states are encountered such that the agent cannot
perform the next operator, it will call Algorithm \ref{algo:learn}
(\learn). Otherwise, the agent continues with the next operator until all
the operators in a plan are complete. If the goal conditions are satisfied
(line 16), the algorithm returns success. Otherwise, it turns to \learn.

\begin{algorithm}[t]
	\small
	\caption{\solve($\symdp$, $\mstate$, $\reachfluentstates$,
		$\planfluentstates$, $\threshold$, $\lflag$, $\learners$)}
		\begin{algorithmic}[1]
			\REQUIRE{$\symdp$: Integrated Planning Task}
			\REQUIRE{$\mstate$: Initial MDP state from which to plan}
			\REQUIRE{$\reachfluentstates$: Set of reachable fluent states}
			\REQUIRE{$\planfluentstates$: Set of plannable fluent states}
			\REQUIRE{$\threshold$: Value threshold parameter}
			\REQUIRE{$\lflag$: $\true$ if this algorithm was called from $\learn$}
			\REQUIRE{$\learners$: A set of learners}
			\STATE{$\fluentstate \gets \task.\detector(\mstate)$}
			\IF {$\symdp.\goal \subseteq \fluentstate$}
				\RETURN {$\lflag, \learners$}
			\ENDIF
			\STATE {$\plan, \mathit{visitedNodes} \gets \owfs(\symdp.\task, \fluentstate)$}
			\STATE{$\reachfluentstates.\mbox{add}(\mathit{visitedNodes})$}
			\IF {$\plan \neq \emptyset$}
				\STATE{$\planfluentstates.\mbox{add}(\mbox{all }\mathit{visitedNodes}\mbox{ along $\plan$})$}
				\FOR{operator $\operator_i$ in $\plan$}
				\STATE{$\mstate \gets \execute(\symdp.\executor(\operator_i), \mstate)$}
				\STATE{$\fluentstate \gets
				\symdp.\detector(\mstate)$}
					\IF {$\precon(\operator_{i+1}) 	\nsubseteq \fluentstate$}
						\RETURN {\learn($\symdp$, $\mstate$, $\reachfluentstates$, $\planfluentstates$, $\threshold$, $\learners$)} 
					\ENDIF
				\ENDFOR
				\IF {$\symdp.\goal \subseteq \fluentstate$}
				\RETURN {$\lflag, \learners$}
				\ELSE
					\RETURN {\learn($\symdp$, $\mstate$, $\reachfluentstates$, $\planfluentstates$, $\threshold$, $\learners$)} 
				\ENDIF
			\ELSE
			\RETURN {\learn($\symdp$, $\mstate$, $\reachfluentstates$, $\planfluentstates$, $\threshold$, $\learners$)} 
			\ENDIF
		\end{algorithmic}
		\label{algo:solve}
\end{algorithm}

%
%


\subsection{Learning Operator Policies} 
\label{sec:learning}
Broadly, in Algorithm \ref{algo:learn} (\learn), the agent follows an
exploration policy $\explorationpolicy$, e.g., $\epsilon$-greedy, to explore the environment, spawning RL
agents which attempt to construct policies to reach particular partial fluent
states from which the goal $\goal$ is reachable by planning.  These partial
fluent states correspond to operator effects. For each such set of effects, the
system attempts to find sets of preconditions from which that operator can
consistently achieve high value; when one such set of preconditions is found,
the policy along with the corresponding set of preconditions and effects is
used to define a new operator and its corresponding executor, which are added
to the IPT.

\begin{algorithm}[b]
	\small
	\caption{$\learn(\symdp, \mstate, \reachfluentstates, \planfluentstates, \threshold, \learners)$}
	\begin{algorithmic}[1]
	\REQUIRE {$\symdp$: Integrated Planning Task}
	\REQUIRE{$\mstate$: Initial MDP state}
	\REQUIRE{$\reachfluentstates$: Set of reachable fluent states}
	\REQUIRE{$\planfluentstates$: Set of plannable fluent states}
	\REQUIRE {$\threshold$: Value threshold}
	\REQUIRE{$\learners$: set of learners}
	\STATE {$done \gets \false$}
	\STATE{$\fluentstate \gets \task.\detector(\mstate)$}
	\WHILE {$\lnot done$}
		\STATE {$\action \sim \explorationpolicy(\cdot\mid\mstate)$}
		\STATE {$\mstate' \sim \task.\transitionprob(\cdot \mid \mstate, \action)$}
		\STATE {$\fluentstate \gets
		\symdp.\detector(\mstate')$}
		\IF {$\fluentstate \supseteq \partialfluentstate$ for some $\partialfluentstate \in \planfluentstates$ }
		\STATE {$done \gets \true$}
		\ELSE
		\STATE {$\plan, \mathit{visitedNodes} \gets
			\owfs(\symdp.\task, \fluentstate)$}
		
		\IF {$\plan \neq \emptyset$}
			\STATE {$\subgoal \gets \goal$}
			\FOR {$\operator_i$ in reversed$(\plan)$}
				\STATE {$\subgoal \gets \regressfunction(\subgoal, \operator_i)$}
				\STATE{$\planfluentstates.\mbox{add}(\subgoal)$}
				\STATE{$\offlinelearner_{\subgoal} \gets \mbox{spawnLearner}(\subgoal)$}
				\STATE{$\learners \gets \learners \cup \{ \offlinelearner_{\subgoal} \}$}
			\ENDFOR
			\STATE{$done \gets \true$}
		\ENDIF
		\ENDIF
	\STATE{$\onlinelearner.\mbox{train}(\mstate, \action, \mathbf{1}_{done=\true}, \mstate')$}
	\FOR{$\offlinelearner_{\subgoal} \in \learners$}
	\STATE{$\offlinelearner_{\subgoal}.\mbox{train}(\mstate, \action, \mathbf{1}_{\fluentstate \supseteq \subgoal}, \mstate')$}
	\ENDFOR
	\STATE{$\mstate \gets \mstate'$}
	\ENDWHILE
	\FOR {$\offlinelearner_{\subgoal} \in \learners$}
	\FOR {$\partialfluentstate_{pre} \in \genprecon(\offlinelearner_{\subgoal},
		\reachfluentstates, \threshold)$}
	\STATE {$\operator^\star \gets \langle \partialfluentstate_{pre}, \subgoal, \emptyset \rangle$}
	\STATE {$\execstar \gets \mbox{makeExecutor}(\offlinelearner_{\subgoal}, \operator^\star)$}
	\STATE{$\symdp.\mbox{addOperator}(\operator^*, \execstar)$}
	\ENDFOR
	\ENDFOR
	\RETURN {\solve($\symdp$, $\mstate$, $\reachfluentstates$, $\planfluentstates$, $\threshold$, $\true$, $\learners$)}
\end{algorithmic}
		\label{algo:learn}
\end{algorithm}

Algorithm receives as input, among other things, a set of learners $\learners$ (which may grow during execution, as described below). $\learners$ includes an exploration agent $\onlinelearner$ with corresponding policy $\explorationpolicy$ (initialized in Algorithm~\ref{algo:overall}).  $\learners$ also contains a set of
offline ``subgoal'' learners, which initially is empty. 

In each time step, \learn~ executes an action $\action$ according to its exploration policy $\explorationpolicy$, with resulting MDP state $\mstate'$ and corresponding fluent state $\fluentstate$ (lines 4-6). The agent will
attempt to check if $\fluentstate$ is a fluent state from which it can plan to the goal. First the agent checks if $\fluentstate$ is already \textit{known} to be a state from which it can plan to the goal ($\fluentstate \in \planfluentstates$; line 7).  If not, the agent attempts to plan from $\fluentstate$ to the goal (line 10). If there is a plan, then the agent regresses each operator in the plan in reverse order ($\operator_n, \cdots, \operator_1$).  As described in Section~\ref{sec:planning}, after regressing through $o_i$, the resulting fluent state $\partialfluentstate_{i-1}$ is the most general possible partial fluent state (i.e., containing the least possible fluents) such that executing $\langle \operator_i, \cdots, \operator_n\rangle$ from $\partialfluentstate_{i-1}$ results in $\goal$; this makes each $\partialfluentstate_{i-1}$ a prime candidate for some new operator $\operator^\star$'s effects. That is, assuming $\static(\operator)=\emptyset$, $\partialfluentstate_{i-1}$ guarantees that $\langle \operator_i, \cdots, \operator_n \rangle$ is a plan to $\goal$, while allowing the corresponding executor to terminate in the largest possible set of fluent states. Each such partial fluent state $\subgoal$ is chosen as a subgoal, for which a new learner $\offlinelearner_{\subgoal}$ is spawned and added to $\learners$ (lines 12-18).  $\subgoal$ is also added to the set of ``plannable'' fluent states $\planfluentstates$.

Each subgoal learner $\offlinelearner_{\subgoal}$ is trained each time step using as reward the indicator function $\mathbf{1}_{\fluentstate \supseteq \subgoal}$, which returns $1$ if that learner's subgoal is satisfied by $\mstate'$ and $0$ otherwise (lines 23-25).

While in this case the exploration learner $\onlinelearner$ may be conceived as an RL agent whose reward function is $1$ whenever \textit{any} such subgoal is achieved (line 22) and whose exploration policy $\explorationpolicy$ is  $\epsilon$-greedy, nothing prevents it from being defined differently as a random agent or even symbolic
learner as discussed in Section \ref{sec:related_work}.

Upon reaching a state from which a plan to the goal exists, the agent stops exploring. For each of its subgoal learners $\offlinelearner_{\subgoal} \in \learners$, it attempts to construct sets of preconditions (characterized as a partial fluent state $\partialfluentstate_{pre}$) from which its policy can consistently achieve the subgoal state $\subgoal$ (see Section~\ref{sec:precon}) (lines 28-34).  If any such precondition sets $\partialfluentstate_{pre}$ exist, the agent constructs the operator $\operator^\star$ such that $\precon(\operator^\star) = \partialfluentstate_{pre}$, $\effect(\operator^\star)=\subgoal$, and without static fluents (all other variables are unknown once the operator is executed; $\static(\operator^\star)=\emptyset$).  The corresponding executor is constructed as $\mbox{makeExecutor}(\offlinelearner_{\subgoal}, \operator^\star) = \langle \execinitstar, \execpolicystar, \exectermstar \rangle$, where 
\begin{align*}
\execinitstar &= \{ \mstate' \in \mstates : \task.\detector(\mstate') \supseteq \precon(\operator^\star) \} \\
\execpolicystar(\action \mid \mstate_{init}, \mstate) &= \begin{cases}
1 &\mbox{if }\action=\argmax\limits_{\action' \in \task.\mdp.\actions} \offlinelearner_{\subgoal}.q(\mstate, \action) \\
0 & \mbox{otherwise}
\end{cases} \\
\exectermstar(\mstate_{init}, \mstate) &= \begin{cases}
1 &\mbox{if }\task.\detector(\mstate) \supseteq \subgoal \\
0 &\mbox{otherwise}.
\end{cases}
\end{align*}
Note that while the definition of the Operator Discovery Problem allows the constructed operators' executors to depend on $\mstate_{init}$, in practice the operators are ordinary options.  Options are sufficient, \textit{provided the operators being constructed have no static fluents.}\footnote{This also indicates why we used our open-world planning formalism: in closed-world planning, all fluents not in the effects are static, requiring executors which depend on $\mstate_{init}$. By specifying static fluents for each operator, we can leverage handmade operators which (largely) satisfy the closed-world assumption while allowing the operators returned by SPOTTER to be open-world.} Constructing operators with static fluents is a topic for future work.

Constructed operators and their corresponding executors are added to the IPT (line 32), which ideally becomes plannable as a result, solving the Operator Discovery Problem. Because the agent is currently in a state from which it can plan to the goal, control is passed back to \solve~(with $\lflag$ set to $\true$ because this episode required learning).

\subsubsection{Learning operator policies in GridWorld}

The puzzle shown in Figure~\ref{fig:screenshot} presents a Stretch-IPT  for
which a solution exists in terms of the MDP-level actions, but no planning solution exists.
Thus, SPOTTER enters \learn, and initially moves around randomly (the exploration policy has not yet received a reward). Eventually, in the course of random action, the agent moves the ball out of the way (Algorithm \learn, line 11). From this state, the action plan
\begin{multline*} \plan = goToObj(agent, key);  pickUp(agent, key); \\ goToObj(agent, door);  useKey(agent, door)\end{multline*} achieves the goal $\goal = \{open(door)\}$. Regressing through this plan from the goal state (Algorithm \learn, lines 13-17), the agent identifies the subgoal
\begin{multline*} \subgoal = \{ nextToFacing(agent, ball), \\handsFree(agent), inRoom(agent, key), \\ inRoom(agent, door), locked(door), \\  \neg holding(agent, key), \neg blocked(door) \}\end{multline*} (as well as other subgoals corresponding to the suffixes of $\plan$), and constructs a corresponding subgoal learner, $\offlinelearner_{\subgoal}$ which uses RL to learn a policy which consistently achieves $\subgoal$.

\subsection{Generating Preconditions}
\label{sec:precon}

\begin{algorithm}[b]\centering
	\small
	\caption{$\genprecon(\offlinelearner, \reachfluentstates,
	\threshold)$}
	\begin{algorithmic}[1]
		\REQUIRE {$\offlinelearner$: Learner along with Q-tables}
		\REQUIRE {$\reachfluentstates$: Fluent states reachable from the
		initial state}
		\REQUIRE {$\threshold$: Value threshold}
		\STATE {$\abovethresholdfluentstates \gets \emptyset$}
		\STATE {$\mathit{queue} \gets \emptyset$}
		\FOR {$\fluentstate$ in $\reachfluentstates$}
		\IF {$\mbox{value}(\fluentstate) > \threshold$}
		\STATE {$\mathit{queue}.\mbox{add}(\fluentstate)$}
		\STATE {$\abovethresholdfluentstates.\mbox{add}(\fluentstate)$}
		\ENDIF
		\ENDFOR
		\STATE {$\beenfluentstates \gets
		\mbox{getFluentStatesFromQ}(\offlinelearner)$}
		\WHILE {$\mathit{queue}$}
		\STATE {$\node \gets \mathit{queue}.\mbox{pop}()$}
		\FOR {$\anotherfluentstate$ in $\beenfluentstates$}
		\STATE {$\common \gets \node \cap \anotherfluentstate$}
		\IF {$\mbox{value}(\common) > \threshold \land \common \mbox{
			not in } \abovethresholdfluentstates$}
		\STATE {$\mathit{queue}.\mbox{add}(\common)$}
		\STATE {$\abovethresholdfluentstates.\mbox{add}(\common)$}
		\ENDIF
		\ENDFOR
		\ENDWHILE
		\RETURN {$\abovethresholdfluentstates$}
		\end{algorithmic}
		\label{algo:genprecon}
\end{algorithm}


The prior algorithms allow an agent to plan in the symbolic space, and when
stuck explore a subsymbolic space with RL learners. We noted that each learner
is connected to a particular subgoal -- from which the agent can generate
a symbolic plan -- which in turn, maps onto the effects of a potential new
operator. What remains is to define the preconditions of such an operator.
Algorithm \ref{algo:genprecon} (\genprecon) incrementally generates increasingly general
sets of preconditions for each learner. 

\genprecon~ begins with initializing a set of
above-threshold fluent states $\abovethresholdfluentstates$  to empty. This set
will represent the output of the algorithm, which in turn is essentially a form
of graph search over the space of possible sets of preconditions. More
specifically, \genprecon~ first adds all the fluent states in the
set of reachable fluent states $\reachfluentstates$ to the search queue and to
$\abovethresholdfluentstates$ (lines 3-7). The learner
$\offlinelearner$ has visited states in the MDP and has been updating its
Q-table as part of \learn~ (see line 24 in \learn). We probe this Q-table and extract the
fluent states that the agent has visited and populate a set of ``been'' fluent states
$\beenfluentstates$ (line 9). While there are fluent states (or partial fluent
states) in the queue, a set of ``successor'' nodes is computed for each one (node)  by capturing the set of fluents common to both the node and a particular ``been'' fluent state $\fluentstate'$, for each $\fluentstate' \in \beenfluentstates$.  If the average value (according to $\offlinelearner$'s $q$ function) of all states satisfying $\common$ is above the threshold $\threshold$, then the above-threshold
common partial fluent state $\common$ is added to
$\abovethresholdfluentstates$. The idea here is to only allow sets of preconditions which are guaranteed reachable by the planner (generalizations of elements of $\reachfluentstates$), and to compute the set of all such sets of preconditions for which the agent can consistently achieve the learner's subgoal $\subgoal$ (where for the purposes of this paper, a set of preconditions $\partialfluentstate_{pre}$ ``consistently achieves'' $\subgoal$ if the average value of all MDP states satisfying $\partialfluentstate_{pre}$ is above the threshold $\threshold$). \genprecon~ can be terminated at any
time during the while loop, and will yield zero or more plausible preconditions for a particular
learner-operator. If terminated before any nodes have been expanded, \genprecon~ will simply yield elements from
$\reachfluentstates$. However, allowing this algorithm to run longer will yield
more general preconditions, i.e., those with fewer fluents. 


\subsubsection{Precondition generation in GridWorld}

Returning to the puzzle in Figure~\ref{fig:screenshot}, the precondition
generation algorithm uses the value function for the subgoal learner
$\offlinelearner_{\subgoal}$ to determine a conjunction of fluents such that
the MDP states satisfying those fluents have a high average state value (say,
$> 0.9$), and which is plannable from the environment's start
state.\footnote{The preconditions produced are too lengthy to include in this
paper, but can be found in the supplementary material.} This conjunction
specifies the new operator's preconditions. The operator is added to the
planning domain, and this augmented domain can be solved using planning alone,
and can be used in other plans to achieve other goals in the environment.


\section{Experiments}


We evaluate SPOTTER on MiniGrid \cite{gym_minigrid}, a platform based on
procedurally generated 2D gridworld environments populated with objects
-- agent, balls, doors and keys. The agent can navigate the world, manipulate objects, and move between rooms.

To establish the fact that (1) operator discovery is possible, and (2) that the
symbolic operators discovered by SPOTTER are useful for performing new tasks in
the environment, we structure our evaluation into three puzzles that the agent
must solve in sequence, much like three levels in puzzle video games
(Fig.~\ref{fig:screenshot}). The environment in each level consists of two rooms with
a locked door separating them.  The agent and the key to the room are randomly
placed in the leftmost room.  In puzzle 1, the agent's task is to pickup the
key, and use it to open the door. The episode terminates, and the agent
receives a reward inversely proportional to the number of elapsed time steps,
when the door is open.  In puzzle 2 the agent's goal is again to open the door,
but a ball is placed directly in front of the door, which the agent must pick
up and move elsewhere before picking up the key (this is the running example). Puzzle 3 is identical to
puzzle 2, except that the agent's goal is now not to open the door, but to go
to a goal location (green square) in the far corner of the rightmost room. The high-level planning domain and low-level RL actions are as defined for the running example.

The evaluation is structured so that for almost all initial conditions in
puzzle 1, SPOTTER's planner can
produce a plan that can be successfully executed to reach the goal state.  In
puzzle 2, the door is blocked by the ball, and the agent has no operator
representing ``move the ball out of the way'' (and no combination of existing
operators can safely achieve this effect given the agent's planning domain).
The agent must discover such an operator. Finally,
puzzle 3 is designed to test whether the learned operator can be used in the
execution of different goals in the same environment.

\begin{figure}[t]
	\centering
	\scalebox{0.5}
	{\hspace*{-1cm}\includegraphics[width=1\textwidth]{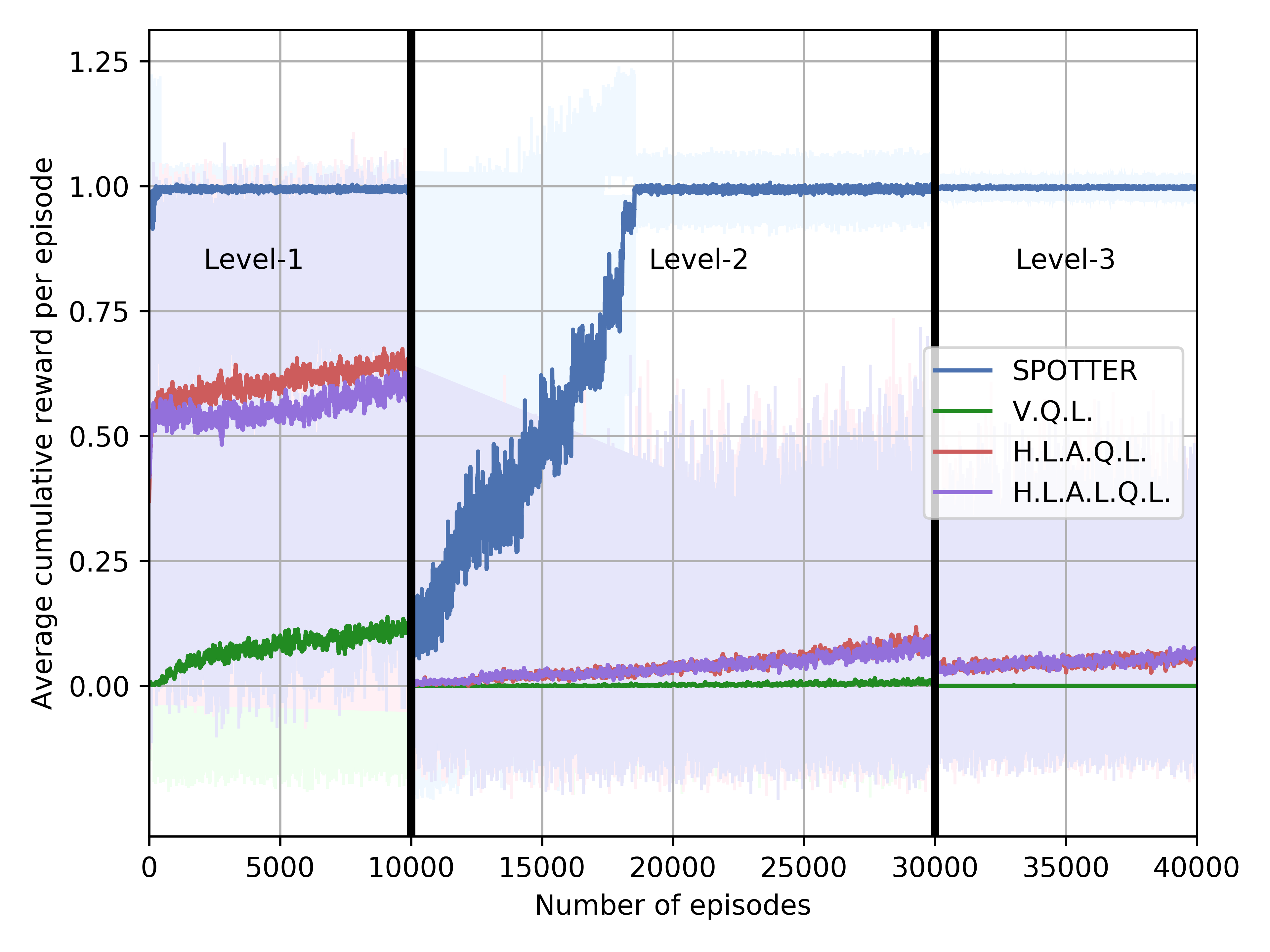}}
	\caption{Experimental performance across three tasks. We report mean
		cumulative reward (and standard deviation in lighter color) obtained by
		our approach (SPOTTER) and three baseline algorithms: tabular Q-learning over primitive actions (VQL), tabular Q-learning with primitive and high-level action executors (HLAQL), and HLAQL with q-updates
		trickling down from HLAs to primitives (HLALQL).}
	\label{fig:results}
\end{figure}

Figure~\ref{fig:results} shows the average results of running SPOTTER 10 times on puzzles 1 through 3.  The algorithm was allowed to retain
any operators/executors discovered between puzzles 2 and 3. In each case we employed a
constant learning rate $\alpha = 0.1$, discount factor $\gamma=0.99$, and an
$\epsilon$-greedy exploration policy with decaying exploration constant
$\epsilon$ beginning at $0.9$ and decaying towards $0.05$. $\epsilon$ is decayed exponentially using the formula
$\epsilon(t) = \epsilon_{min} + (\epsilon_{max} - \epsilon_{min})e^{-\lambda t}$,
where $\lambda = -log(0.01)/N$, $N$ being the maximum number of episodes. The value threshold was set to $\threshold=0.9$. We ran for 10,000 
episodes on puzzle 1, 20,000 episodes on puzzle 2, and 10,000 episodes on
puzzle 3. 

The experimental results show that SPOTTER achieved higher overall rewards than
the baselines in the given time frame and did so more
quickly.\footnote{Code implementing SPOTTER and the baselines along with
	experiments will be made available post-review.} Crucially, the
agent learned the missing operator for moving the blue ball out of the way in
Level 2, and was immediately able to use this operator in Level 3. This is
demonstrated both by the fact that the agent did not experience any drop in
performance when transitioning to Level 3 and also we know from running the
experiment that the agent did not enter \learn~ or \genprecon~ in Level
3. It is important to note that the baselines did converge at around 800,000 -
1,000,000 episodes, significantly later than SPOTTER.\footnote{In the supplementary material, we provide the learned operator
	described in PDDL, learning curves for the baselines over 2,000,000
	episodes, and videos showing SPOTTER's integrated planning and learning.}. 

The results suggest that SPOTTER significantly outperformed comparable
baselines. The HLAQL and HLALQL baselines have their action space augmented
with the same high-level executors provided to SPOTTER. SPOTTER does not use any function approximation techniques and the
exploration and subgoal learners are tabular Q-learners themselves. Accordingly, we
did not compare against any deep RL baselines. We also did not compare transfer learning and curriculum learning approaches as these
approaches do not handle cases where new representations need
to be learned from the environment. 


\begin{figure*}[t]
	\centering
	\begin{subfigure}[t]{.32\textwidth}
		\centering
		\includegraphics[width=1\textwidth]{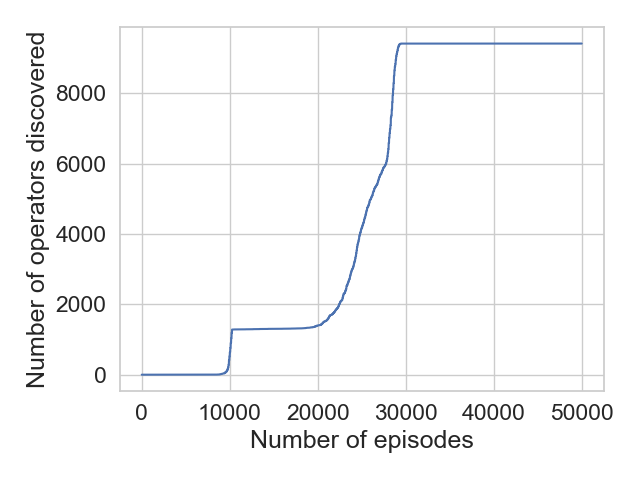}
		\caption{The number of unique sets of preconditions discovered for which average value is above threshold increases as SPOTTER is allowed to continue exploring.}
		\label{fig:precondops}
	\end{subfigure} \hfill
	\begin{subfigure}[t]{.32\textwidth}
		\centering
		\includegraphics[width=1\textwidth]{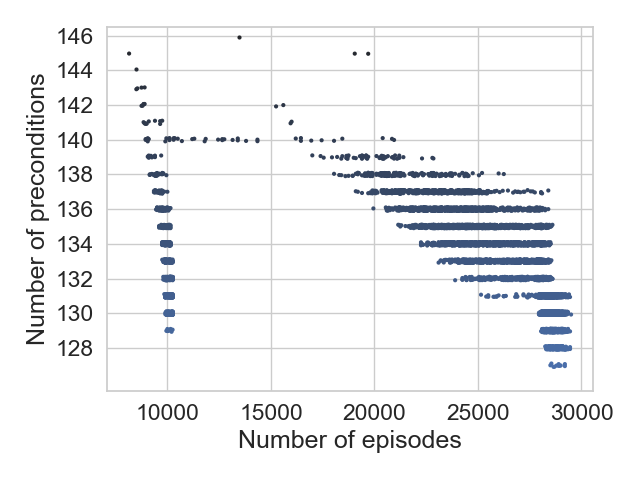}
		\caption{Generation of operators with decreasing numbers of preconditions proceeds in ``waves'' as the value estimates of additional MDP states converge.}
		\label{fig:precondscatter}
	\end{subfigure} \hfill
	\begin{subfigure}[t]{.32\textwidth}
		\centering
		\includegraphics[width=1\textwidth]{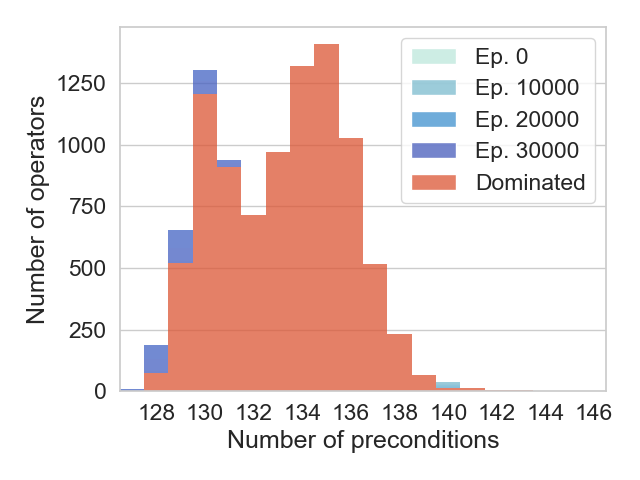}
		\caption{By episode 30,000, almost all operators have been dominated (replaced by superior operators), and almost all non-dominated operators were created late in the run.}
		\label{fig:precondhist}
	\end{subfigure}
	\caption{Results of precondition generalization experiment for a single subgoal learner on puzzle 2.}
	\label{fig:precond}
\end{figure*}

\subsection{Experiment 2}

Ordinarily, as soon as SPOTTER discovers an operator exceeding the value threshold, that operator is incorporated into the planning agent's model. We dispensed with this assumption and ran SPOTTER on puzzle 2 for 50,000 episodes, allowing the system to continue learning operator policies and generating preconditions throughout this time. Every 50 episodes, SPOTTER logged the new operators it had created. (Operators were not logged if their preconditions were specifications of preconditions for which an operator had already been discovered.) Figure~\ref{fig:precond} shows the results for one particular operator learner (i.e., each of the output operators has the same postconditions and the same underlying policy, but has a unique set of preconditions). As Figure~\ref{fig:precondops} indicates, by episode 29,500 this learner discovered 9,049 unique operators; no further operators were discovered after this episode.

Figures~\ref{fig:precondscatter} and ~\ref{fig:precondhist} demonstrate that with additional exploration, the agent can construct more general operators. Recall that an operator with a set of preconditions is accepted whenever the average value of all MDP states satisfying those preconditions is greater than the value threshold (in this experiment, $0.9$). As the values of additional MDP states increase past the threshold, more general operators (with fewer preconditions) cross this threshold. Figure~\ref{fig:precondscatter} plots, for each operator discovered, the episode in which it was logged and its total number of preconditions. Note that there are several ``waves'' of precondition generalization.  Beginning with the discovery of the first viable set of preconditions, as additional states cross the threshold there is a rapid discovery of new operators with additional preconditions. Eventually (a little after 10,000 episodes), the existing operators are sufficiently general that many rarely-seen MDP states would have to be thoroughly explored before more general preconditions can be found. For a while, any new operators discovered (while they are not merely specifications of existing operators) have a larger number of preconditions.  This ultimately culminates in a ``second wave'' in which enough MDP states have been explored that more general operators can be produced, ultimately surpassing the first wave.

Figure~\ref{fig:precondhist} shows that operators created later not only have fewer preconditions than earlier operators, but \textit{dominate} earlier operators (in that the preconditions of the dominated operator are a strict superset of the preconditions of the dominating operator).  Orange bars represent dominated operators, blue bars those for which a superior operator has not yet been found. Relatively few non-dominated operators persist by step 30,000, and nearly all that do were created in the last few thousand episodes, suggesting that running SPOTTER for longer before incorporating operators allows the construction of strictly better (more general) operators.\footnote{An animated version of this chart, showing how operators are constructed and then dominated as the agent continues exploring, appears in the supplementary material.}

\section{Discussion}
\label{sec:discussion}

%
The SPOTTER architecture assumes the existence of a planning domain with
operators which correspond to executors which are more or less correct. It does
\textit{not} assume that these executors are reward-optimal. Further, some
tasks can be more efficiently performed if they are not first split into
subgoals.  Thus, the performance of raw RL systems 
eventually overtakes that of SPOTTER on any particular environment.  This is not
a serious flaw -- SPOTTER also produces knowledge that can be more easily
applied to perform other tasks in the same environment.

%
%
The environments used to test SPOTTER (puzzles 1 through 3) are deterministic.
In stochastic environments, it is often
difficult to design planning domains where operators
have guaranteed effects. While SPOTTER can
handle stochastic environments, it would need more robust metrics for
assessing operator confidence.
%
%

Future work could also emphasize adapting this work to high-dimensional
continuous state and action spaces using deep
reinforcement learning.  ``Symbolizing'' such spaces can be difficult, and in
particular such work would have to rethink how to generate candidate
preconditions, since the existing approach enumerates over all states, which
clearly would not work in deep RL domains. 
%
%

The key advantage of SPOTTER is that the agent can produce operators that can
potentially be applied to other tasks in the same environment. Because these
operators' executors are policies (here, policies over finite, atomic
MDPs), they do not generalize particularly well to environments with different
dynamics or unseen start states (e.g., an operator learned in puzzle 2 could
not possibly be applied in puzzle 3 if the door was moved up or down by even
one cell). While function approximation could be helpful to
solving this problem, an ideal approach might be a form of program
synthesis \cite{solar2009sketching}, in which the agent learns
\textit{programs} that can
be applied regardless of environment.

Furthermore, in this work, we manually sequenced the three tasks to elicit the discovery of operators that would be useful in
the final task environment. A possible avenue for future work would be
to develop automated task sequencing methods (i.e., curriculum
learning \cite{narvekar2020curriculum}) as to improve performance in downstream tasks.

\section{Related Work}
\label{sec:related_work}

In this section, we review related work that overlaps with various aspects of
the proposed integrated planning and learning literature. 

\subsection{Learning Symbolic Action Models}

Early work in the symbolic AI literature explored how agents can adjust and
improve their symbolic models through experimentation. Gill proposed a method for
learning by experimentation in which the agent can improve its
domain knowledge by finding missing operators \cite{gil1994learning}. The agent is able design experiments at the symbolic level based on
observing the symbolic fluent states and comparing against an operator's
preconditions and effects. Other approaches (to name a few:
\cite{shen1989learning,shen1989rule,mitchell1986explanation,gizzi2019creative,joshi2012abstract,talamadupula2010planning}),
comprising a significant body of literature, have explored recovering from
planning failures, refining domain models, and open-world planning. 

These approaches do not handle
the synthesis of the underlying implementation of an operator as we have
proposed. That is, they synthesize new operators, but no executors.  That
said, the rich techniques offered by these approaches can be useful
in integrating into our online learner. As we describe in Section~\ref{sec:learning}, the
online learner follows an $\epsilon$-greedy exploration policy. Future work will explore how our online learner could be
extended to conduct symbolically-guided experiments as 
these approaches suggest.  

More recently, there has been a growing body of literature exploring how domain
models can be learned from action traces
\cite{arora2018review,jimenez2012review,zimmerman2003learning,hogg2010learning}.
The ARMS system learns PDDL planning operators from examples. The examples consist of successful fluent
state and operator pairs corresponding to a sequence of transitions in the
symbolic domain \cite{yang2007learning}.  Subsequent work has explored how
domains can be learned with partial knowledge of successful traces
\cite{aineto2019learning,cresswell2013acquiring}, and with
neural networks capable of approximating from partial traces
\cite{xiao2019representation} and learning models from pixels
\cite{asai2017classical,dittadi2018learning}. 

In the RL context, there has been recent work in learning symbolic
concepts \cite{konidaris2018skills} from low-level actions. Specifically,  
Konidaris et al. assume the agent has available to it a set of high-level actions, couched in the options framework for hierarchical reinforcement
learning. The agent must learn a symbolic planning domain with operators and
their preconditions and effects. This approach to integrating planning and
learning is, in a sense, a reverse of our approach. While their use
case is an agent that has high-level actions but no symbolic representation, ours assumes that we have (through hypothesizing from the backward
search) most of the symbolic scaffolding, but need to learn the policies themselves. 


\subsection{Learning Action Executors}
There has been a tradition of research in leveraging planning domain knowledge and symbolically provided constraints to
improve the sample efficiency of RL systems. Grzes et al. use a potential
function (as a difference between source and destination fluent states) to
shape rewards of a Q-learner \cite{grzes2008plan}. While aligned with our own
targeted Q-learners, their approach requires the existence of symbolic plans,
which our agent does not possess. A related approach (PLANQ) combines STRIPS
planning with Q-learning to allow an agent to converge faster to an optimal
policy \cite{grounds2005combining}. Planning with Partially Specified Behaviors (PPSB) is based on PLANQ,
and takes as input symbolic domain information and produces a Q-values for
each of the high-level operators \cite{aguas2016planning}. Like Grzes et al.,
PLANQ and PPSB assume the existence of a plan, or at least a plannable
task. Other recent approaches have extended these ideas to incorporate using
partial plans  \cite{illanes2020symbolic}. The approach proffered by Illanes et
al. improves the underlying learner's sample efficiency, but can also be provided with symbolic goals. However,
their approach assumes the agent has a complete (partial order) plan for its
environment, and can use that plan to construct a reward function for learning
options to accomplish a task. One key difference between their work and
ours is that we construct our own operators, whereas
they use a set of existing operators to learn policies.

Generally, while these methods learn action implementations, they assume the
existence of a successful plan or operator definition. In the proposed
framework, the agent has neither and must hypothesize operators and their corresponding executors.

RL techniques have also been used to improve the quality of symbolic plans. The
DARLING framework uses symbolic planning and reasoning to constrain 
exploration and reduce the search space, and uses RL to
improve the quality of the plan produced \cite{leonetti2016synthesis}. While our approach shares the common
goal of reducing the brittleness of planning domains, they do not modify the planning model.  The PEORL framework works to choose a plan that maximizes
a reward function, thereby improving the quality of the plan using RL \cite{yang2018peorl}. More recently Lyu et al. propose a framework
(SDRL) for generalizing the PEORL framework with intrinsic goals and
integration with a deep reinforcement learning (DRL) machinery \cite{lyu2019sdrl}.
However, neither PEORL nor SDRL synthesizes new operators or learns planning domains.  Our work
also differs from the majority of model-based RL approaches  \cite{ng2019incremental,winder2019planning} in that we are interested in
STRIPS-like explicit operators that are useful in symbolic planning, and
thereby transferable to a wide range of tasks within a domain.

%

\section{Conclusion}

Automatically synthesizing new operators during task performance is a crucial
capability needed for symbolic planning systems. As we have examined here, such
a capability requires a deep integration between planning and learning, one
that benefits from leveraging RL to fill in missing connections between
symbolic states. While exploring the MDP state space, SPOTTER can identify
states from which symbolic planning is possible, use this to identify subgoals
for which operators can be synthesized, and learn policies for these operators
by RL. SPOTTER thus allows a symbolic planner to explore previously unreachable
states, synthesize new operators, and accomplish new goals.


\section{Acknowledgements}

This work was funded in part by NSF grant IIS-2044786 and DARPA grant
W911NF-20-2-0006.

\bibliographystyle{ACM-Reference-Format} 
\bibliography{refs}  


\end{document}


\title{SPOTTER: Supplementary Material}

\maketitle
\tableofcontents
\newpage
\section{Extended Experiments with Baselines}

Below are figures showing experimental results for our baseline algorithms over
2,000,000 episodes. 

\begin{figure}[H]
	\centering
	\includegraphics[width=5in]{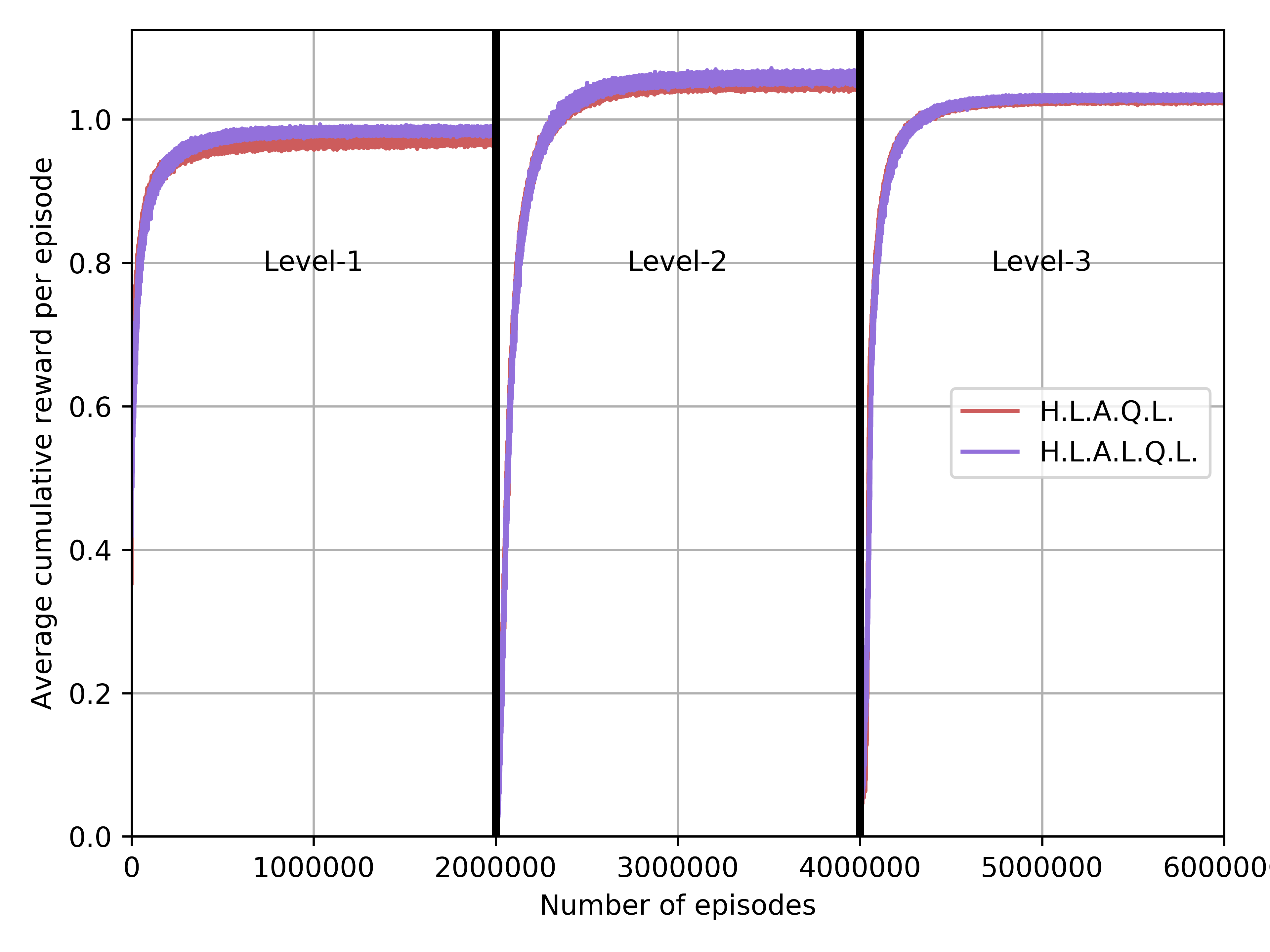}
	\caption{Baselines up to 2,000,000 episodes normalized to maximum reward
	obtained by SPOTTER. In the long run, the baselines perform better than
	SPOTTER, however, they take much longer and do not learn transferable
	knowledge as effectively as SPOTTER.}
	\label{fig:baselines2M}
\end{figure}

\begin{figure}[H]
	\centering
	\includegraphics[width=5in]{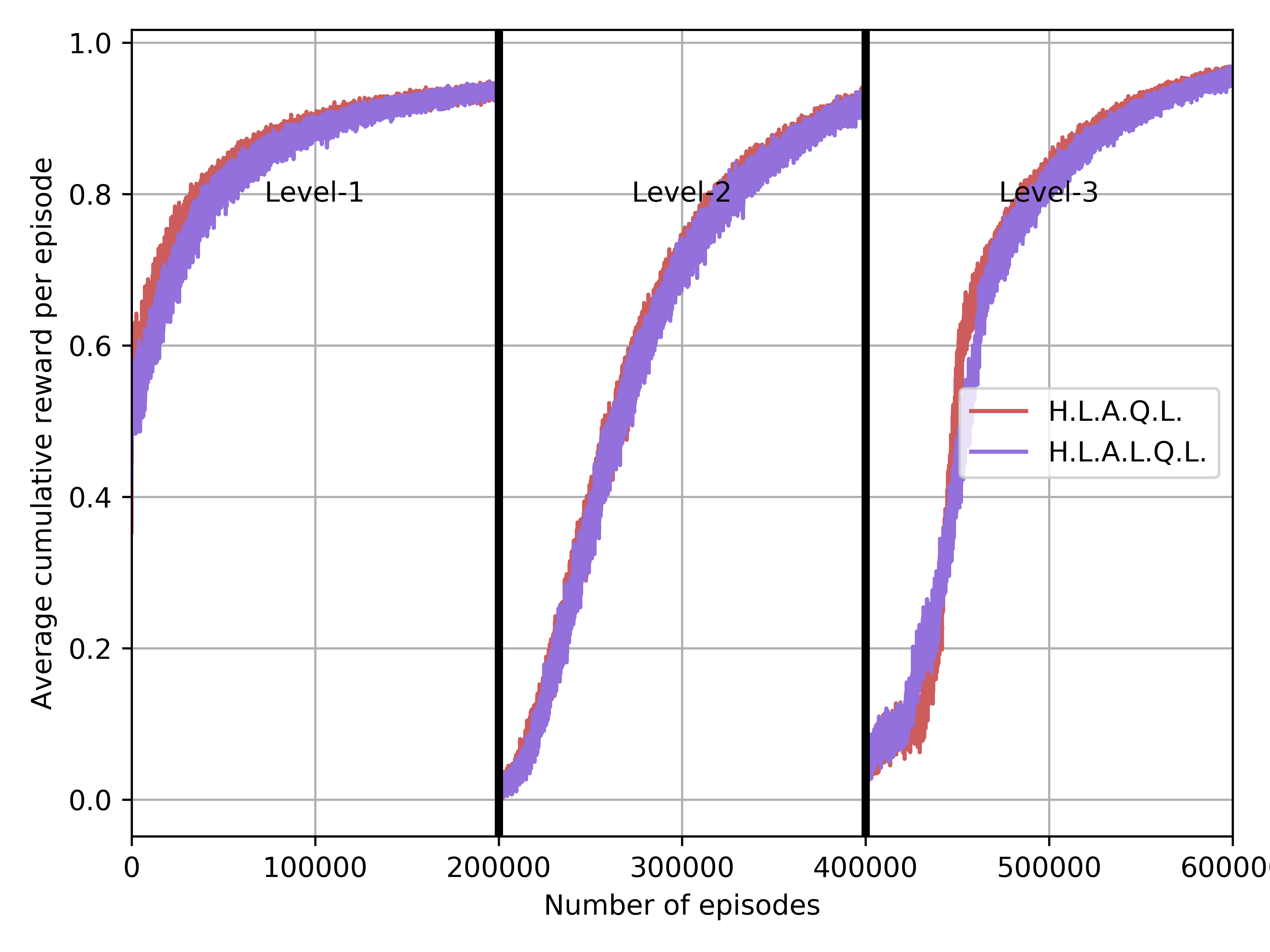}
	\caption{Baselines up to 200,000 episodes  normalized to maximum reward
	obtained by SPOTTER.}
	\label{fig:baselines2M}
\end{figure}

\begin{figure}[H]
	\centering
	\includegraphics[width=5in]{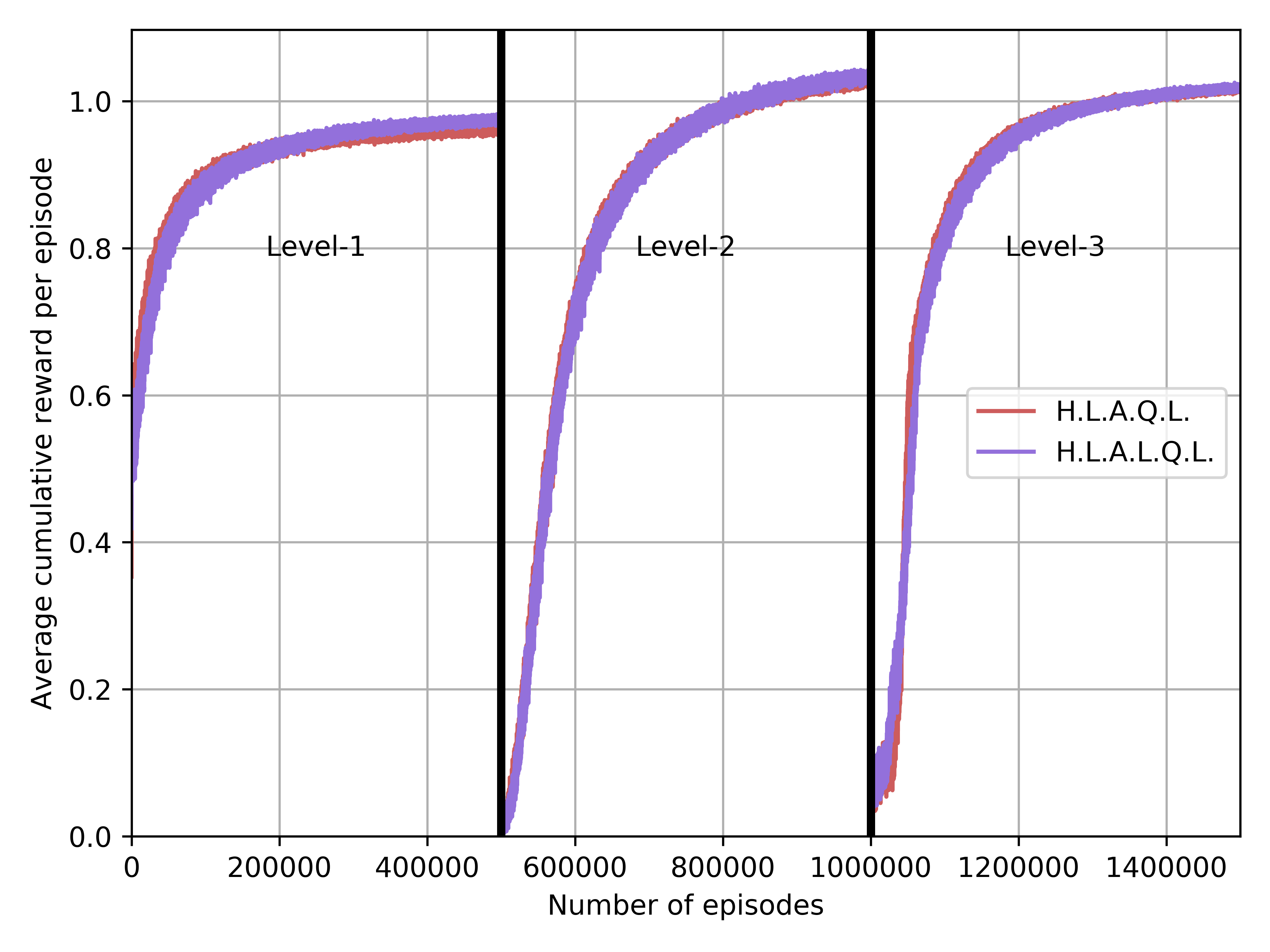}
	\caption{Baselines up to 500,000 episodes  normalized to maximum reward
	obtained by SPOTTER.}
	\label{fig:baselines2M}
\end{figure}

\section{Videos}
Attached in the supplementary materials package are several videos, showing
specific runs of SPOTTER through the environment. Beginning  with level 1,
which SPOTTER can solve entirely by planning (i.e., no need for learning), then
proceeding to learning in Level 2. In Level 2, SPOTTER at first needs to
explore (episode 0), then when it hits a plannable state, can transition to
planning the rest of the way to the goal, and then when it finds the operator,
it can complete Level 2 entirely without any RL and only with planning. One
promising aspect of SPOTTER is that in Level 3, we show that it can simply use
the learned operator in a plan, without instantiating a new learning session,
even though the goals have changed substantially. 

\begin{itemize}
	\item SPOTTER Level 1 (planning)
	\item SPOTTER Level 2 (episode 0)
	\item SPOTTER Level 2 (hit plannable state)
	\item SPOTTER Level 2 (found operator)
	\item SPOTTER Level 3 (plan with learned operator)
\end{itemize}

We also provide a video for precondition generalization in which we show, over
time, how the distribution of preconditions evolves. The orange bars show that
more general preconditions have superceded less general ones (blue).

\section{Open World PDDL}

Below is the original PDDL domain given to SPOTTER for the 2-room-blocked
minigrid environment. 

\footnotesize
\begin{lstlisting}{language=Lisp}
(define (domain gridworld_abstract)

  (:requirements :strips :typing )

  (:types thing room - object
          agent physobject - thing
          door nondoor - physobject
          notgraspable graspable - nondoor
          wall floor goal lava - notgraspable
          key ball box - graspable)

  (:predicates
           (nexttofacing ?a - agent ?thing - physobject)
           (open ?t - door)
           (closed ?t - door)
           (locked ?t - door)
           (holding ?a - agent ?t - graspable)
           (handsfree ?a - agent)
           (obstructed ?a - agent)
           (blocked ?t - door)
           (atgoal ?a - agent ?g - goal)
           (inroom ?a - agent ?g - physobject)
           )

  (:action pickup
   :parameters (?a - agent ?thing - graspable)
   :precondition (and (handsfree ?a) (nexttofacing ?a ?thing) (not (holding ?a ?thing)))
   :unknown (and)
   :effect (and (not (handsfree ?a))
                (not (nexttofacing ?a ?thing))
                (not (obstructed ?a))
                (holding ?a ?thing))
  )

  (:action putdown
   :parameters (?a - agent ?thing - graspable)
   :precondition (and (not (obstructed ?a)) (holding ?a ?thing))
   :unknown (and (blocked *))
   :effect (and (not (holding ?a ?thing))
                (handsfree ?a)
                (nexttofacing ?a ?thing)
                (obstructed ?a)
                ))

  (:action gotoobj1
   :parameters (?a - agent ?thing - graspable)
   :precondition (and (not (holding ?a ?thing)) (not (obstructed ?a)) (inroom ?a ?thing))
   :unknown (and)
   :effect (and (obstructed ?a)
                (nexttofacing ?a ?thing)))

  (:action gotoobj2
   :parameters (?a - agent ?thing - notgraspable)
   :precondition (and (not (obstructed ?a)) (inroom ?a ?thing))
   :unknown (and)
   :effect (and (obstructed ?a)
                (nexttofacing ?a ?thing)))

  (:action gotoobj3
   :parameters (?a - agent ?thing - graspable ?obstruction - thing)
   :precondition (and (not (holding ?a ?thing)) (nexttofacing ?a ?obstruction) 
   (inroom ?a ?thing))
   :unknown (and)
   :effect (and (nexttofacing ?a ?thing)
                (not (nexttofacing ?a ?obstruction))))

  (:action gotoobj4
   :parameters (?a - agent ?thing - notgraspable ?obstruction - thing)
   :precondition (and (nexttofacing ?a ?obstruction) (inroom ?a ?thing))
   :unknown (and)
   :effect (and (nexttofacing ?a ?thing)
                (not (nexttofacing ?a ?obstruction))))

  (:action usekey
   :parameters (?a - agent ?key - key ?door - door)
   :precondition (and (nexttofacing ?a ?door) (holding ?a ?key) (locked ?door))
   :unknown (and)
   :effect (and (open ?door) (not (closed ?door)) (not (locked ?door))))

  (:action opendoor
   :parameters (?a - agent ?door - door)
   :precondition (and (nexttofacing ?a ?door) (closed ?door))
   :unknown (and)
   :effect (and (open ?door) (not (closed ?door)) (not (locked ?door))))

  (:action stepinto
   :parameters (?a - agent ?g - goal)
   :precondition (and (nexttofacing ?a ?g))
   :unknown (and (nexttofacing ?a *))
   :effect (and (not (nexttofacing ?a ?g)) (atgoal ?a ?g)))

  (:action gotodoor1
   :parameters (?a - agent ?thing - door)
   :precondition (and (not (obstructed ?a)) (not (blocked ?thing)) (inroom ?a ?thing))
   :unknown (and)
   :effect (and (obstructed ?a)
                (nexttofacing ?a ?thing)))

  (:action gotodoor2
   :parameters (?a - agent ?thing - door ?obstruction - thing)
   :precondition (and (nexttofacing ?a ?obstruction) (not (blocked ?thing)) 
   (inroom ?a ?thing))
   :unknown (and)
   :effect (and (nexttofacing ?a ?thing)
                (not (nexttofacing ?a ?obstruction))))

  (:action enterroomof
   :parameters (?a - agent ?d - door ?g - physobject)
   :precondition (and (not (blocked ?d)) (nexttofacing ?a ?d) (open ?d))
   :unknown (and)
   :effect (and (inroom ?a ?g) (not (nexttofacing ?a ?d)) (not (obstructed ?a))))

)
\end{lstlisting}
\normalsize

\section{Discovered Operator}

Below is an example of an operator discovered by SPOTTER. 

\begin{lstlisting}
+PR: (inroom agent obj0013)
  +PR: (inroom agent key)
  +PR: (inroom agent obj0026)
  +PR: (inroom agent obj0005)
  +PR: (inroom agent obj0011)
  +PR: (inroom agent obj0028)
  +PR: (locked door)
  +PR: (inroom agent obj0009)
  +PR: (inroom agent obj0023)
  +PR: (inroom agent obj0001)
  +PR: (inroom agent obj0007)
  +PR: (inroom agent obj0002)
  +PR: (inroom agent obj0020)
  +PR: (holding agent key)
  +PR: (inroom agent obj0025)
  +PR: (inroom agent obj0016)
  +PR: (inroom agent obj0006)
  +PR: (nexttofacing agent obj0015)
  +PR: (inroom agent obj0012)
  +PR: (inroom agent obj0017)
  +PR: (inroom agent obj0027)
  +PR: (inroom agent obj0003)
  +PR: (inroom agent obj0000)
  +PR: (inroom agent obj0022)
  +PR: (inroom agent obj0015)
  +PR: (inroom agent obj0031)
  +PR: (inroom agent obj0008)
  +PR: (inroom agent obj0019)
  +PR: (inroom agent obj0004)
  +PR: (inroom agent obj0029)
  +PR: (inroom agent obj0021)
  +PR: (inroom agent ball)
  +PR: (inroom agent obj0030)
  +PR: (inroom agent obj0014)
  +PR: (inroom agent obj0024)
  +PR: (inroom agent door)
  +PR: (inroom agent obj0018)
  +PR: (inroom agent obj0010)
  +PR: (obstructed agent)
  -PR: (inroom agent obj0038)
  -PR: (inroom agent obj0058)
  -PR: (inroom agent obj0062)
  -PR: (nexttofacing agent obj0004)
  -PR: (nexttofacing agent obj0023)
  -PR: (nexttofacing agent obj0032)
  -PR: (nexttofacing agent obj0014)
  -PR: (inroom agent obj0047)
  -PR: (nexttofacing agent obj0029)
  -PR: (nexttofacing agent obj0046)
  -PR: (nexttofacing agent obj0016)
  -PR: (inroom agent obj0059)
  -PR: (inroom agent obj0042)
  -PR: (nexttofacing agent obj0052)
  -PR: (nexttofacing agent obj0049)
  -PR: (nexttofacing agent obj0006)
  -PR: (nexttofacing agent obj0033)
  -PR: (nexttofacing agent obj0003)
  -PR: (inroom agent obj0054)
  -PR: (inroom agent obj0048)
  -PR: (inroom agent obj0055)
  -PR: (nexttofacing agent obj0027)
  -PR: (nexttofacing agent obj0063)
  -PR: (inroom agent obj0057)
  -PR: (nexttofacing agent obj0013)
  -PR: (nexttofacing agent obj0048)
  -PR: (inroom agent obj0053)
  -PR: (nexttofacing agent goal)
  -PR: (holding agent ball)
  -PR: (nexttofacing agent obj0012)
  -PR: (inroom agent obj0033)
  -PR: (inroom agent obj0040)
  -PR: (nexttofacing agent obj0051)
  -PR: (inroom agent obj0045)
  -PR: (inroom agent obj0039)
  -PR: (nexttofacing agent obj0041)
  -PR: (nexttofacing agent obj0022)
  -PR: (inroom agent obj0036)
  -PR: (inroom agent obj0051)
  -PR: (nexttofacing agent obj0058)
  -PR: (handsfree agent)
  -PR: (nexttofacing agent obj0040)
  -PR: (inroom agent obj0035)
  -PR: (nexttofacing agent obj0002)
  -PR: (inroom agent obj0050)
  -PR: (nexttofacing agent obj0026)
  -PR: (inroom agent obj0041)
  -PR: (nexttofacing agent obj0028)
  -PR: (atgoal agent goal)
  -PR: (nexttofacing agent obj0054)
  -PR: (nexttofacing agent obj0045)
  -PR: (nexttofacing agent obj0059)
  -PR: (inroom agent obj0056)
  -PR: (inroom agent obj0046)
  -PR: (inroom agent obj0034)
  -PR: (nexttofacing agent ball)
  -PR: (nexttofacing agent obj0009)
  -PR: (nexttofacing agent obj0010)
  -PR: (inroom agent obj0044)
  -PR: (nexttofacing agent obj0008)
  -PR: (nexttofacing agent obj0025)
  -PR: (nexttofacing agent obj0030)
  -PR: (nexttofacing agent obj0024)
  -PR: (nexttofacing agent obj0011)
  -PR: (nexttofacing agent obj0061)
  -PR: (nexttofacing agent key)
  -PR: (nexttofacing agent obj0005)
  -PR: (inroom agent goal)
  -PR: (closed door)
  -PR: (nexttofacing agent obj0053)
  -PR: (nexttofacing agent agent)
  -PR: (inroom agent obj0052)
  -PR: (nexttofacing agent obj0044)
  -PR: (open door)
  -PR: (nexttofacing agent obj0043)
  -PR: (inroom agent obj0037)
  -PR: (nexttofacing agent obj0019)
  -PR: (nexttofacing agent obj0021)
  -PR: (nexttofacing agent door)
  -PR: (nexttofacing agent obj0050)
  -PR: (nexttofacing agent obj0056)
  -PR: (inroom agent obj0063)
  -PR: (nexttofacing agent obj0047)
  -PR: (nexttofacing agent obj0020)
  -PR: (nexttofacing agent obj0039)
  -PR: (nexttofacing agent obj0007)
  -PR: (nexttofacing agent obj0031)
  -PR: (inroom agent obj0032)
  -PR: (nexttofacing agent obj0000)
  -PR: (nexttofacing agent obj0034)
  -PR: (nexttofacing agent obj0062)
  -PR: (inroom agent obj0060)
  -PR: (inroom agent obj0043)
  -PR: (nexttofacing agent obj0060)
  -PR: (nexttofacing agent obj0038)
  -PR: (inroom agent obj0049)
  -PR: (nexttofacing agent obj0055)
  -PR: (nexttofacing agent obj0001)
  -PR: (nexttofacing agent obj0017)
  -PR: (nexttofacing agent obj0042)
  -PR: (inroom agent obj0061)
  -PR: (nexttofacing agent obj0037)
  -PR: (nexttofacing agent obj0035)
  -PR: (nexttofacing agent obj0036)
  -PR: (nexttofacing agent obj0018)
  -PR: (nexttofacing agent obj0057)
  ADD: (nexttofacing agent ball)
  ADD: (handsfree agent)
  ADD: (inroom agent key)
  ADD: (inroom agent door)
  ADD: (locked door)
  DEL: (holding agent key)
  DEL: (blocked door)
\end{lstlisting}

%
%
%